\documentclass[11pt]{article}

\usepackage[final]{acl}
\usepackage{times}
\usepackage{latexsym}
\usepackage[T1]{fontenc}
\usepackage[utf8]{inputenc}
\usepackage{microtype}
\usepackage{inconsolata}
\usepackage{graphicx}
\usepackage{booktabs}
\usepackage{array}
\usepackage{tabularx}
\usepackage{multirow}
\usepackage{makecell}
\usepackage{amsmath}
\usepackage{amssymb}
\usepackage{xcolor}
\usepackage{url}
\usepackage{algorithm}
\usepackage{algorithmic}
\usepackage{enumitem}

\newcommand{\method}{\textbf{LatBridge}}

\newcommand{\jiajun}[1]{{\color{black}#1}}
\title{LatBridge: Efficient and Stable Model Replacement in \\Latent-Communication Multi-Agent Systems}
\author{
  Jiajun Zhu$^{1}$\thanks{First author: \texttt{jzhucq@connect.ust.hk}}, Haobo Li$^{1}$, Zixin Chen$^{1}$, Huamin Qu$^{1}$, Wenshuo Zhang$^{1}$\thanks{Corresponding author: \texttt{wzhangeb@connect.ust.hk}} \\
  $^{1}$The Hong Kong University of Science and Technology
}

\begin{document}
\maketitle

\begin{abstract}
Multi-agent systems rely on communication to complete tasks, and latent communication passes continuous representations to reduce text generation and re-encoding.
\jiajun{However, multi-agent systems must reconfigure agent models, whereas existing latent-communication methods are built for fixed model combinations, so their interfaces may fail after replacement.}
Our pilot study confirms this interface failure and identifies the requirement for interface reuse after model replacement.
We present \method{}, which connects agent models through a shared Bridge and separate interfaces under a shared latent-communication protocol.
After agent model replacement, \method{} reuses unchanged interfaces, selects matching low-rank interfaces, and combines them through the Bridge, training only missing ones with all models frozen.
\jiajun{The Bridge processes source representations in the basic mode and constructs target input without autoregressive generation in the accelerated mode.}
\jiajun{Experiments support effective communication after model replacement, with mean basic-mode accuracies of 14.8\% on MATH and 27.3\% on ALFWorld next-action prediction across eight small-model configurations.}
\jiajun{Analyses examine the factors supporting task accuracy and communication efficiency.}
\end{abstract}
\section{Introduction}

Multi-agent systems organize language models as agents with different roles and use communication to exchange information, divide tasks, and make collaborative decisions~\citep{li2023camel,hong2024metagpt,wu2024autogen}.
Most existing systems communicate through text, but generating and re-encoding intermediate text increases communication time and converts continuous internal representations into discrete text.
To reduce these limitations, previous studies have explored latent communication, which exchanges continuous representations such as embeddings~\citep{pham2024cipher}, activations~\citep{ramesh2025activations}, hidden states~\citep{du2026interlat}, and key-value caches~\citep{shi2026kvcomm}.
At the same time, multi-agent systems need to select models according to task and resource requirements~\citep{yue2025masrouter} and adjust communication topology for different forms of collaboration~\citep{li2024sparse,shen2025topology}.
These changes may require replacing the model assigned to an agent.
However, existing latent-communication methods mainly build communication interfaces for fixed sending and receiving models~\citep{du2026interlat,zou2025latentmas,fu2025cachetocache}.
It remains unclear whether communication interfaces trained by these methods can continue to pass useful information after the model assigned to an agent is replaced.

\begin{figure}[t]
    \centering
    \includegraphics[width=0.96\columnwidth]{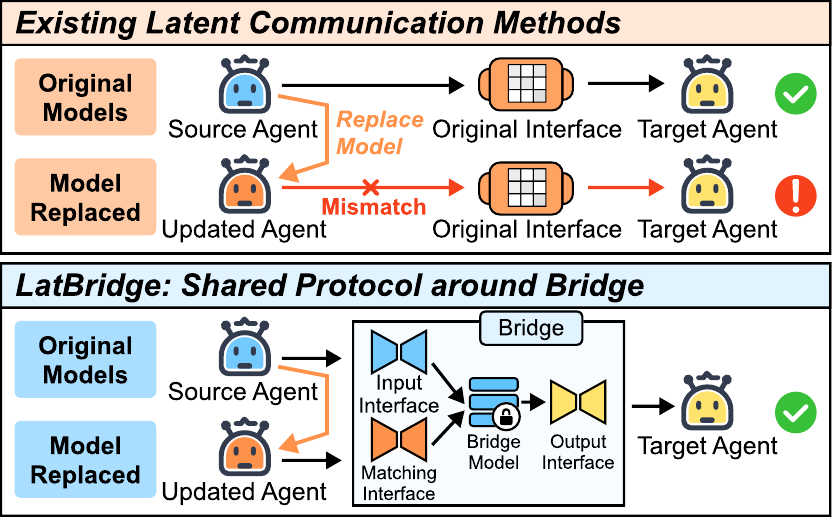}
    \caption{Agent model replacement in latent communication.
    \jiajun{Interfaces in latent-communication methods can fail after replacement.}
    \jiajun{\method{} keeps the Bridge model fixed, reuses the unchanged interface, and selects or trains a matching interface for the new model.}}
    \label{fig:concept}
    \vspace{-0.4cm}
\end{figure}

To test whether the original communication interface remains usable after model replacement, we conduct a pilot study based on Interlat~\citep{du2026interlat}.
The results show that replacing the model assigned to an agent without retraining the interface substantially reduces the original communication gain, indicating that an interface trained for the original model combination cannot be directly applied to a new endpoint model.
This result motivates a shared communication protocol for all agents, implemented through a shared communication component that divides the model-pair-dependent representation mapping across its two sides.
With this protocol, replacing a model changes only its connection to this component, allowing the interface on the unchanged side to be reused and a new adjacent interface to be trained only when no matching interface exists.
However, if this component uses a complete language model to generate and process communicated content token by token, this additional step increases communication latency and makes communication time depend on the generated sequence length.
We therefore need a latent-communication method that reduces both the interface retraining required after model replacement and the processing latency introduced by the shared communication component.

To this end, we present \method{} for efficient and stable model replacement in multi-agent systems.
\autoref{fig:concept} contrasts interface mismatch after model replacement with interface reuse through the Bridge.
Structurally, \method{} implements the shared communication component as the Bridge and connects the sending source agent and the receiving target agent through an input interface, a fixed Bridge model, and an output interface.
When a model is replaced, the system reuses the unchanged interface and looks for a stored interface matching the new model.
The two interfaces are connected through the fixed Bridge model.
Only missing interfaces are trained, with all models frozen.
The input and output interfaces learn representation mappings between adjacent models in separate training stages.
Trained interfaces are stored by their adjacent models for use in the current and future matching configurations.
Once the required interfaces are available, the system communicates through the input interface, complete frozen Bridge model, and output interface, preserving source-dependent communication.
This procedure is called the basic mode.
To reduce the latency caused by token-by-token Bridge generation, \method{} prepares a vocabulary mapping and a corresponding output interface in advance, then uses the Bridge tokenizer and embedding lookup to construct target-side continuous input directly during communication.
This procedure is called the accelerated mode.
In the accelerated mode, the system bypasses autoregressive generation with the complete Bridge model, reducing communication processing time.

To evaluate \method{}, we conduct model-replacement, communication-efficiency, ablation, and representation analyses at two model scales and across different endpoint-model configurations.
\jiajun{As existing methods do not provide a suitable direct baseline for agent model replacement, we use controlled experiments, with Interlat as an accuracy reference.}
\jiajun{With matched interfaces, mean basic-mode accuracy across eight small-model configurations is 14.8\% on MATH and 27.3\% on ALFWorld next-action prediction.}
\jiajun{The scaling experiments show effective communication across the tested configurations.}
\jiajun{Ablation and representation analyses examine how interfaces, training, and communication content support task accuracy after replacement.}
\jiajun{The accelerated mode reduces Bridge processing latency while maintaining or improving average task accuracy.}
Complete experimental settings, results, and analyses are provided in \autoref{sec:experiments}.

Our contributions are threefold.

\begin{itemize}
    \item We identify interface failure after model replacement as an unresolved problem in \jiajun{latent-communication methods}, and use a pilot study to show that the original interface \jiajun{cannot be directly reused} after model replacement.

    \item We present \method{}, which implements a shared communication protocol through a communication component shared by all agents to support interface reuse after model replacement and avoids token-by-token Bridge generation in the accelerated mode.

    \item We evaluate \method{} in terms of task accuracy, communication efficiency, component roles, and representation space.
    \jiajun{The results show that matched interfaces support task accuracy after agent model replacement, while the accelerated mode improves communication efficiency.}
\end{itemize}
\section{Related Work}
\label{sec:related-work}

Multi-agent systems coordinate role-specialized agents and have developed from dialogue-based collaboration to structured workflows~\citep{li2023camel,hong2024metagpt,wu2024autogen}.
Because text-based communication adds token generation and re-encoding overhead and may discard internal information, latent-communication methods pass embeddings~\citep{pham2024cipher}, activations~\citep{ramesh2025activations}, hidden states~\citep{du2026interlat}, or key-value caches~\citep{shi2026kvcomm}, with recent work further studying joint latent collaboration~\citep{zou2025latentmas}, cache transfer~\citep{fu2025cachetocache}, learned messages~\citep{zheng2025thought}, and cross-model conversion~\citep{liu2026xkv}.
Meanwhile, model assignment~\citep{yue2025masrouter}, communication topology~\citep{li2024sparse}, model diversity~\citep{chen2024reconcile}, and interaction design~\citep{liang2024debate} affect both performance and cost, creating a practical need to replace agent models while preserving existing task divisions and communication relations.

However, existing latent-communication methods mainly build communication interfaces for fixed sending and receiving models, leaving interface reuse after model replacement unresolved.
Drawing on model stitching~\citep{bansal2021stitching} and low-rank adaptation~\citep{hu2022lora}, \method{} establishes a shared communication protocol.
This allows \method{} to extend latent communication to changing agent-model configurations, with unaffected and matching interfaces reused after model replacement.

\section{Background and Motivation}
\label{sec:motivation}

This section examines the difficulty of reusing interfaces from existing latent-communication methods after model replacement and motivates our design.
\autoref{sec:motivation-problem} states this interface reuse problem, and \autoref{sec:motivation-pilot} examines it through a pilot study.

\subsection{Problem Statement}
\label{sec:motivation-problem}

\jiajun{As reviewed in \autoref{sec:related-work}, latent communication enables agents to exchange continuous representations, while multi-agent systems may need to replace agent models.}
\jiajun{However, interfaces in latent-communication methods are trained for fixed sending and receiving models.}
\jiajun{Model replacement can change the representations these interfaces receive or produce, but prior work has not tested whether the original interfaces remain usable.}

This leaves an unexamined problem:
\emph{After the model assigned to an agent in a multi-agent system is replaced, can a latent-communication interface trained for the original model combination be reused directly and continue to pass task-relevant information to the target agent?}

To answer this question, we model a directed communication relation in a multi-agent system as one-way communication between two models.
A multi-agent system can be represented as a graph whose nodes are agents and whose directed edges are communication relations.
Applying this model to each directed communication relation extends the analysis to multi-agent systems with different sizes and communication topologies.
The sending model corresponds to the source agent, and the receiving model corresponds to the target agent, while both agents retain their task roles in the system.
The next subsection examines this problem through a pilot study of model replacement.

\subsection{Pilot Study}
\label{sec:motivation-pilot}

\begin{table}[t]
\centering
\scriptsize
\setlength{\tabcolsep}{4pt}
\renewcommand{\arraystretch}{0.96}
\begin{tabular}{@{}llccc@{}}
\toprule
Task & Target & Original & No communication & Replacement \\
\midrule
\multirow{2}{*}{MATH}
& Qwen  & $12.00\!\pm\!0.00$ & $0.50\!\pm\!0.00$ & $0.67\!\pm\!0.24$ \\
& Llama & $5.00\!\pm\!0.00$  & $0.00\!\pm\!0.00$ & $0.50\!\pm\!0.41$ \\
\addlinespace[1pt]
\multirow{2}{*}{ALFWorld}
& Qwen  & $23.63\!\pm\!2.53$ & $0.25\!\pm\!0.36$ & $0.00\!\pm\!0.00$ \\
& Llama & $19.41\!\pm\!4.09$ & $0.00\!\pm\!0.00$ & $0.00\!\pm\!0.00$ \\
\bottomrule
\end{tabular}
\caption{Pilot study of source-agent model replacement based on Interlat~\citep{du2026interlat}.
Accuracies are percentages reported as mean $\pm$ standard deviation.}
\label{tab:pilot}
\vspace{-0.4cm}
\end{table}

\textbf{Experimental setting.}
We use Interlat as a representative latent-communication method~\citep{du2026interlat}.
We use Qwen2.5-0.5B-Instruct~\citep{yang2024qwen25} and Llama-3.2-1B-Instruct~\citep{grattafiori2024llama3} as the source- and target-agent models and evaluate their communication on MATH~\citep{hendrycks2021math} and ALFWorld~\citep{shridhar2021alfworld}.
We compare the original model combination, source-agent model replacement with the target-agent model and original interface fixed, and no communication; complete condition definitions and evaluation details are provided in \autoref{sec:pilot-details}.

\textbf{Results.}
Under the original model combinations, latent communication provides clear performance gains on both tasks.
As shown in \autoref{tab:pilot}, source-agent model replacement retains only $1.5$--$10\%$ of the original communication gain on MATH and none of the original gain on ALFWorld.

\textbf{Analysis.}
Because the target-agent model, task setting, and original interface remain fixed, the decline shows that the replacement model's representations are incompatible with the original interface.
The original interface therefore cannot be directly reused after model replacement.
Supporting model replacement requires separating communication from a fixed endpoint-model combination.

\textbf{Motivation.}
Existing latent-communication methods rely on interfaces trained for fixed model pairs, motivating separate representation mappings so that replacing one endpoint leaves the mapping on the other side unchanged.
A natural solution is to establish a shared communication protocol implemented through a shared communication component used by all agents in a multi-agent system.
For each directed communication relation, separate source-side and target-side interfaces divide the model-pair-dependent conversion across the two sides of this component.
When an agent's model is replaced, only its connection to this component changes, so the interface on the unchanged side can be reused and a new interface is trained only when no matching interface exists.
This shared protocol forms the basis of \method{} in \autoref{sec:latbridge}.

\section{LatBridge}
\label{sec:latbridge}

\begin{figure*}[t]
    \centering
    \includegraphics[width=0.96\textwidth]{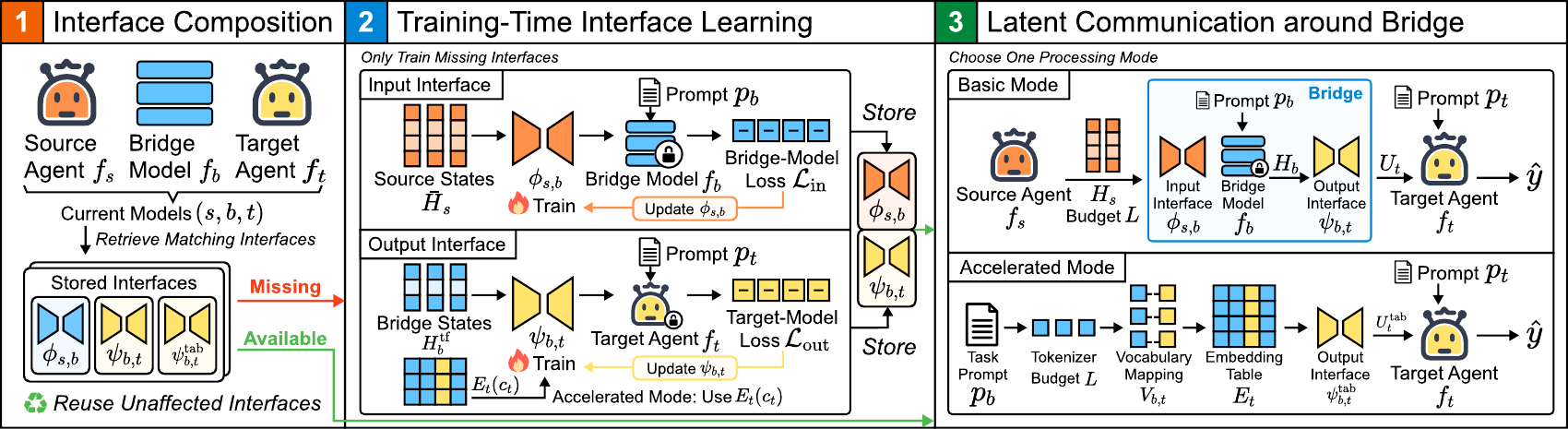}
    \caption{\jiajun{Workflow of \method{}.
    (1) Matching interfaces are retrieved and composed after agent model replacement.
    (2) Only missing interfaces are trained and stored.
    (3) Latent communication around the Bridge uses either the basic or accelerated mode.}}
    \label{fig:overview}
    \vspace{-0.4cm}
\end{figure*}

Building on the motivation in \autoref{sec:motivation}, \method{} supports agent model replacement through a shared communication component and reusable interfaces, while reducing Bridge processing time.
\autoref{sec:latbridge-overview} gives an overview of the method and its workflow after model replacement.
\autoref{sec:latbridge-interface} describes the interfaces and their use after model replacement.
\autoref{sec:latbridge-efficient} explains how to speed up communication under the new model configuration.

\subsection{Overview}
\label{sec:latbridge-overview}

\jiajun{\method{} supports model replacement through a shared communication protocol implemented by the Bridge.}
\autoref{fig:overview} shows the workflow after agent model replacement: interface retrieval and composition, training and storage of missing interfaces, and latent communication around the Bridge.
The following paragraphs explain these steps and how communication is accelerated.
\jiajun{\autoref{sec:interface-structure} details the interface structure, and \autoref{sec:interface-reuse} explains interface use after model replacement.}
\jiajun{\autoref{sec:latbridge-efficient} describes communication acceleration.}

\textbf{Interface retrieval and composition} finds matching interfaces for the new configuration \((s,b,t)\).
\jiajun{Here, \(s\), \(b\), and \(t\) denote the source-agent, Bridge, and target-agent models, with model functions \(f_s\), \(f_b\), and \(f_t\).}
\jiajun{The Bridge contains an input interface \(\phi_{s,b}\), a frozen Bridge model, and an output interface \(\psi_{b,t}\).}
The input and output interfaces connect the source and target models to the Bridge model, separating the representation conversion into two adjacent mappings.
Replacement changes only one adjacent connection, so the system reuses the unchanged interface and retrieves a matching interface for the new model.
\jiajun{Matching interfaces from other communication relations in the system can be reused.}
Once both interfaces are available, they are composed through the fixed Bridge model.

\textbf{Training missing interfaces} supplies a mapping when no stored interface matches the adjacent model pair, with all models frozen and only the missing interface trained.
\jiajun{The input and output interfaces are trained in separate stages.}
The input interface is trained to let the Bridge model read source-agent representations, while the output interface is trained to let the target-agent model read the Bridge output.
The interface is stored by its adjacent models for reuse and combined with the unchanged interface.

\textbf{Latent communication around the Bridge} transfers source-agent information to the target agent after model replacement.
Let \(x_s\), \(p_b\), \(p_t\), and \(\hat{y}\) denote the source input, Bridge-side task prompt, fixed target instruction, and final output.
The operation \(\mathcal{P}_L\) controls the length of source states \(H_s=f_s(x_s)\) under communication budget \(L\).
The input interface \(\phi_{s,b}\) maps these states into Bridge-model inputs.
Conditioned on these inputs and \(p_b\), \(f_b\) generates a sequence.
The output interface \(\psi_{b,t}\) maps the hidden states at generated positions into target-side continuous input.
The target model \(f_t\) generates \(\hat{y}\) using this input and \(p_t\).

\textbf{Efficient communication} reduces the latency added by token-by-token Bridge generation after model replacement.
\method{} prepares a vocabulary mapping and output interface, then uses embedding lookup to construct target-side continuous input directly from the Bridge-side task prompt without this generation step.
\autoref{sec:latbridge-efficient} explains this procedure and its relation to the complete communication path.

\jiajun{For example, consider replacing an agent's model in a multi-agent system using \method{}.}
\jiajun{For its incoming and outgoing communication, the system selects matching interfaces, reuses the unchanged sides, and trains only missing mappings.}
\jiajun{In the accelerated mode, the target-matched mapping and output interface construct continuous input without autoregressive Bridge generation.}
\jiajun{This limits retraining, supports communication with the new model, and reduces communication time.}

\subsection{Interface Design}
\label{sec:latbridge-interface}

Interface design addresses representation differences after model replacement and allows unaffected interfaces to be reused.
\autoref{sec:interface-structure} describes the Bridge and interface structure, and \autoref{sec:interface-reuse} explains interface retrieval and composition, training, storage, and communication.

\subsubsection[Interface Structure]{Interface Structure}
\label{sec:interface-structure}

The Bridge connects endpoint models through a fixed Bridge model, providing a shared connection point for interface reuse after model replacement.
The input interface maps length-controlled source states \(\bar{H}_s\) into representations the Bridge model can consume.
The output interface maps Bridge hidden states \(H_b\) into target-side representations \(U_t\).
Each interface connects one adjacent model pair, so a single interface does not need to learn a mapping that depends on the complete endpoint-model combination.
The following describes how these interfaces handle representation differences between adjacent models.

Models differ in hidden dimension, value distribution, and internal representation, so their hidden states do not correspond directly.
Following model stitching and cross-model representation research~\citep{bansal2021stitching,huh2024platonic}, \method{} uses separate interfaces to learn mappings between the adjacent model pairs.
For an input representation \(h\), a low-rank interface is written as

\vspace{-0.4cm}
\begin{equation}
g(h)
=
W_0h
+
\frac{\alpha}{r}BA\,\operatorname{Dropout}(h),
\label{eq:low-rank-interface}
\end{equation}

where \(W_0\) is the fixed base mapping, \(A\) and \(B\) are trainable low-rank matrices, \(r\) is the rank, and \(\alpha\) is the scaling coefficient.
This parameterization follows the basic form of low-rank adaptation~\citep{hu2022lora} and limits trainable parameters to the interfaces between frozen models.

\subsubsection[Interface Use After Model Replacement]{Interface Use After Model Replacement}
\label{sec:interface-reuse}
\label{sec:interface-construction}

After agent model replacement, \method{} connects the new configuration while retaining the unchanged interface.
The following describes interface retrieval and composition, training of missing interfaces, storage, and communication.

\textbf{Interface retrieval and composition} identifies reusable and missing interfaces for the new configuration.
The system retrieves matching input and output interfaces from its stored collection.
For configuration \((s,b,t)\), the Bridge maps source-agent hidden states to target-side input representations as

\vspace{-0.4cm}

\begin{equation}
G_{s,b,t}
:=
\psi_{b,t}
\circ f_b
\circ \phi_{s,b}.
\label{eq:interface-composition}
\end{equation}

With the Bridge model fixed, \autoref{eq:interface-composition} limits the change to the interface adjacent to the replaced model.
Replacing the source-agent model requires a matching \(\phi_{s,b}\) while \(\psi_{b,t}\) is reused.
Replacing the target-agent model instead requires a matching \(\psi_{b,t}\) while \(\phi_{s,b}\) is reused.
Stored interfaces, including those from other communication relations, can be reused without retraining when their adjacent models and required processing match.

\textbf{Training missing interfaces} provides mappings unavailable for the new model configuration.
Only the missing interface is trained, with all three models frozen.
For a missing input interface, source-agent hidden states are mapped through \(\phi_{s,b}\) and supplied with the Bridge-side task prompt, training the input interface to support Bridge-model generation.
For a missing output interface, the frozen Bridge model produces teacher-forced task states \(H_b^{\mathrm{tf}}\), which \(\psi_{b,t}\) maps into continuous input for the target-agent model.
Let \(y\) denote the training target sequence, with the two interfaces trained using the following objectives

\vspace{-0.4cm}

\begin{align}
\mathcal{L}_{\mathrm{in}}
&:= -\log p_{f_b}\!\left(y \mid p_b,\phi_{s,b}(\bar{H}_s)\right),
\label{eq:input-interface-loss}\\
\mathcal{L}_{\mathrm{out}}
&:= -\log p_{f_t}\!\left(y \mid p_t,\psi_{b,t}(H_b^{\mathrm{tf}})\right).
\label{eq:output-interface-loss}
\end{align}

Each training stage updates only its corresponding interface, so model replacement does not require retraining the complete communication path.

\textbf{Interface storage} makes trained mappings available for reuse across communication relations and later model replacements.
Interfaces are added to the shared collection and indexed by their adjacent models and, when needed, processing mode.
Complete record fields are provided in \autoref{sec:interface-records}.

\textbf{Latent communication around the Bridge} resumes communication once matching interfaces are available after replacement.
The system follows \autoref{eq:interface-composition}, with all models and interfaces frozen during inference.

\subsection{Efficient Communication}
\label{sec:latbridge-efficient}

Efficient communication reduces Bridge processing time after model replacement.
Once interfaces are available, the full Bridge path, called the basic mode, preserves source-dependent communication but adds token-by-token generation latency.
To avoid this generation, \method{} directly constructs target-side continuous input using a prepared vocabulary mapping and output interface.
This procedure is called the accelerated mode and retains the same target-side input convention.
The following describes the training-time preparation and communication-time processing.

\textbf{Training-time preparation} establishes the vocabulary correspondence and trains the output interface \(\psi^{\mathrm{tab}}_{b,t}\) for each Bridge-model and target-agent-model pair.
Let \(c_b\) denote the length-controlled Bridge-side token sequence.
When the two models share a vocabulary, \(c_t=c_b\).
When they use different vocabularies, the system precomputes a vocabulary mapping \(V_{b,t}\) that converts each Bridge token into one or more target-side tokens, producing \(c_t\).
The corresponding output interface \(\psi^{\mathrm{tab}}_{b,t}\) is trained with the target-model cross-entropy objective in \autoref{eq:output-interface-loss}, using \(E_t(c_t)\) instead of teacher-forced Bridge hidden states.
Here, \(E_t\) is the target-agent model's embedding table.
The trained \(\psi^{\mathrm{tab}}_{b,t}\) follows the storage and reuse procedure in \autoref{sec:interface-reuse}, indexed by the Bridge model, target-agent model, and processing mode.

\textbf{Communication-time processing} constructs target-side continuous input using the prepared mapping and output interface.
The system tokenizes \(p_b\) and applies budget \(L\) to obtain \(c_b\).
It obtains \(c_t\) using the vocabulary correspondence above and queries \(E_t\).
The output interface transforms these embeddings as

\vspace{-0.4cm}

\begin{equation}
U_t^{\mathrm{tab}}
=
\psi^{\mathrm{tab}}_{b,t}
\left(
E_t(c_t)
\right).
\label{eq:accelerated-mode}
\end{equation}

The target agent generates its output from \(p_t\) and \(U_t^{\mathrm{tab}}\).
The budget limits the communication sequence length.
Token selection, padding masks, vocabulary mapping, and caching details are provided in \autoref{sec:accelerated-mode-details}.
\section{Experiments}
\label{sec:experiments}

This section evaluates how \method{} supports efficient and stable agent model replacement through task accuracy after replacement, communication efficiency, and the factors affecting its performance.
We study the following three research questions.

\begin{itemize}[leftmargin=*, labelindent=0pt, itemindent=0pt]
    \item \textbf{RQ1: Task Accuracy.}
    After agent model replacement, can \method{} maintain task accuracy using matched interfaces?

    \item \textbf{RQ2: Communication Efficiency.}
    After agent model replacement, can \method{} reduce processing time while maintaining task accuracy?

    \item \textbf{RQ3: Component and Factor Analysis.}
    How do the interface components, training procedure, communication content, communication budget, and representation space affect the performance of \method{} after model replacement?
\end{itemize}

\autoref{sec:exp-setup}, \autoref{sec:exp-replacement}, \autoref{sec:exp-efficiency}, and \autoref{sec:exp-analysis} describe the shared setup and answer the three research questions, respectively.

\begin{table*}[t]
\centering
\small
\setlength{\tabcolsep}{1.5pt}
\renewcommand{\arraystretch}{0.92}
\begin{tabular*}{\textwidth}{@{\extracolsep{\fill}}l@{\hspace{1.5em}}l@{\hspace{1.5em}}lcccccccc@{}}
\toprule
\multirow{3}{*}{Bridge} & \multirow{3}{*}{Source} & \multirow{3}{*}{Target}
& \multicolumn{4}{c}{Qwen2.5-0.5B / Llama-3.2-1B}
& \multicolumn{4}{c}{Qwen2.5-7B / Llama-3.1-8B} \\
\cmidrule(lr){4-7}\cmidrule(l){8-11}
& & & \multicolumn{2}{c}{\jiajun{Basic mode}} & \multicolumn{2}{c}{\jiajun{Accelerated mode}}
& \multicolumn{2}{c}{\jiajun{Basic mode}} & \multicolumn{2}{c}{\jiajun{Accelerated mode}} \\
\cmidrule(lr){4-5}\cmidrule(lr){6-7}\cmidrule(lr){8-9}\cmidrule(l){10-11}
& & & MATH & ALFWorld & MATH & ALFWorld & MATH & ALFWorld & MATH & ALFWorld \\
\midrule
\multirow{4}{*}{Qwen} & Qwen & Qwen
& 16.0 & 33.6 & 16.5 & 24.6
& 30.0 & 26.9 & 57.0 & 45.5 \\
& Qwen & Llama
& 11.0 & 32.8 & 16.0 & 42.5
& 39.0 & 24.6 & 44.0 & 44.8 \\
& Llama & Qwen
& 14.0 & 31.3 & 16.5 & 24.6
& 28.5 & 30.6 & 57.0 & 45.5 \\
& Llama & Llama
& 18.0 & 21.6 & 16.0 & 42.5
& 38.5 & 27.6 & 44.0 & 44.8 \\
\midrule
\multirow{4}{*}{Llama} & Qwen & Qwen
& 16.5 & 22.4 & 19.5 & 29.9
& 27.5 & 40.3 & 59.0 & 44.8 \\
& Qwen & Llama
& 14.5 & 23.9 & 20.0 & 49.3
& 17.0 & 29.9 & 42.0 & 44.0 \\
& Llama & Qwen
& 12.5 & 26.1 & 19.5 & 29.9
& 29.5 & 23.9 & 59.0 & 44.8 \\
& Llama & Llama
& 16.0 & 26.9 & 20.0 & 49.3
& 19.5 & 20.1 & 42.0 & 44.0 \\
\midrule
\multicolumn{3}{l}{Mean over eight configurations}
& 14.8 & 27.3 & 18.0 & 36.6
& 28.7 & 28.0 & 50.5 & 44.8 \\
\bottomrule
\end{tabular*}
\caption{Task accuracy (\%) across eight model configurations at two model scales, grouped by Bridge model.}
\label{tab:task-accuracy}
\vspace{-0.4cm}
\end{table*}

\subsection{Experimental Setup}
\label{sec:exp-setup}

This section describes the settings shared by all experiments.
Each subsequent section introduces only the additional controls and interventions required for its research question.

\textbf{Models and configurations.}
Our main experiments use Qwen2.5-0.5B-Instruct and Llama-3.2-1B-Instruct, and our scaling experiments use Qwen2.5-7B and Llama-3.1-8B-Instruct~\citep{yang2024qwen25,grattafiori2024llama3}.
At each scale, the source-agent model, Bridge model, and target-agent model are selected from the two model families, resulting in eight configurations.
Agent model replacement changes only the source-agent model or target-agent model, while the Bridge model remains fixed within each comparison.

\textbf{Data and tasks.}
We evaluate MATH using normalized exact match of the final answer~\citep{hendrycks2021math}, and ALFWorld using normalized exact match of the next action~\citep{shridhar2021alfworld}.
For both tasks, the target agent receives only a fixed task instruction and the continuous input produced by the Bridge, without direct access to the original task input.
Dataset splits and scoring details are provided in Appendix~\ref{sec:data-scoring-details}.

\textbf{Interface training.}
All language models remain frozen, and the input and output interfaces are trained in two stages following \autoref{sec:interface-construction}.
The two stages use the losses defined in \autoref{eq:input-interface-loss} and \autoref{eq:output-interface-loss}, respectively, with a communication budget of \(L=128\).
The complete training settings are provided in Appendix~\ref{sec:interface-implementation-details}.

\textbf{Evaluation.}
\jiajun{We use controlled experiments because existing methods do not provide a suitable direct baseline for agent model replacement.}
\jiajun{The Interlat pilot study provides an accuracy reference, while no-communication and ablation conditions assess the contribution of communication and individual components.}
The no-communication condition removes the Bridge output and retains only the fixed task instruction of the target agent.
Communication efficiency is measured using synchronized GPU timing, with the timing protocol provided in Appendix~\ref{sec:timing-appendix}.
The message replacement, communication budget, and representation analyses use trained interfaces without further model training.

\subsection[RQ1: Task Accuracy]{RQ1: Task Accuracy}
\label{sec:exp-replacement}

This section evaluates whether \method{} maintains task accuracy after agent model replacement through interface reuse and missing-interface training.
The results show that it supports all tested configurations at both model scales and in both the basic and accelerated modes.

\textbf{Settings and controls.}
In each comparison, we replace one agent's model while keeping the Bridge model and the other agent's model fixed.
The system reuses the interface on the unchanged side and selects or trains only the interface adjacent to the new model.
The complete replacement operations are provided in Appendix~\ref{sec:replacement-operations}.

\textbf{Results.}
\autoref{tab:task-accuracy} reports the main small-model experiments and the large-model scaling experiments.
Across the four model-scale--task aggregates, average accuracy ranges from 14.8\% to 28.7\% in the basic mode and from 18.0\% to 50.5\% in the accelerated mode.
All large-model configurations in the basic mode achieve nonzero task accuracy, and the accelerated mode performs better in all 16 configuration--task comparisons.

\textbf{Analysis.}
Compared with the substantial degradation caused by retaining the original interface in \autoref{tab:pilot}, matched interfaces restore effective communication under the new agent-model configurations.
Changing the source-agent model changes the basic-mode results, showing that this mode preserves source-dependent information, whereas the accelerated mode provides a more stable target-side continuous input.
These results support interface selection, interface reuse, and missing-interface training as ways for \method{} to maintain task accuracy after model replacement while retaining the interface on the unchanged side.

\subsection{RQ2: Communication Efficiency}
\label{sec:exp-efficiency}

This section evaluates whether \method{} reduces Bridge processing time in the accelerated mode while maintaining task accuracy, once interfaces for the new models are available.

\textbf{Settings.}
We compare the basic and accelerated modes under the shared models, data, interfaces, and communication budget in \autoref{sec:exp-setup}.
Timing covers Bridge processing and output adaptation; the complete boundary and hardware settings are provided in Appendix~\ref{sec:timing-appendix}.

\textbf{Results.}
The aggregate timing results are reported in \autoref{tab:latency}, and the task accuracies are shown in \autoref{tab:task-accuracy}.
Across both tasks and model scales, average communication time ranges from 559.84 ms to 2,256.96 ms in the basic mode and from 0.334 ms to 0.367 ms in the accelerated mode.
The accelerated mode reduces average communication time by approximately \(1{,}669\times\) to \(6{,}149\times\), while maintaining or improving average task accuracy in all four model-scale--task aggregates.

\textbf{Analysis.}
Token-by-token generation with the complete Bridge model accounts for most of the Bridge processing latency.
By avoiding these autoregressive forward passes, \method{} reduces the processing overhead of the shared communication component in the accelerated mode while maintaining or improving average task accuracy.
Thus, matched interfaces support communication after model replacement, and acceleration reduces Bridge processing time during communication.

\subsection{RQ3: Component and Factor Analysis}
\label{sec:exp-analysis}

This section examines which designs in \method{} support effective communication after model replacement and how communication content, budget, and representation differences affect task performance under the new configurations.

\textbf{Module ablations.}
We ablate the interface components and training procedure in the small-model basic-mode setting while keeping the remaining experimental conditions fixed.
As reported in \autoref{tab:ablation}, removing either interface or using coupled output training reduces accuracy on both tasks, while joint optimization has task-dependent effects and updating the Bridge model provides no consistent improvement.
These results show that separate interfaces and decoupled two-stage training support effective communication and provide a basis for adapting the adjacent interface while reusing the unchanged interface after model replacement.
The complete variants, settings, and learned-parameter counts are provided in Appendix~\ref{sec:ablation-details}.

\begin{figure}[t]
  \centering
  \includegraphics[width=\columnwidth]{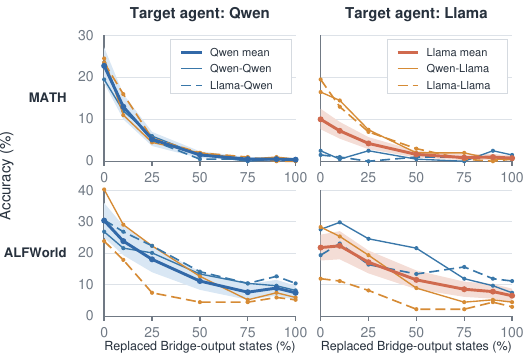}
  \caption{Task accuracy under Bridge-output replacement \jiajun{in the basic mode}.
  \jiajun{Thin curves show model configurations, labeled by source and Bridge models.}
  \jiajun{Thick curves show target-family means with bootstrap 95\% confidence bands.}}
  \label{fig:dose-response}
  \vspace{-0.4cm}
\end{figure}

\textbf{Communication content and budget.}
We replace increasing proportions of the Bridge-output states before the output interface while keeping the models, interfaces, and decoding settings fixed.
As shown in \autoref{fig:dose-response}, accuracy decreases consistently across both tasks and target-agent families as the replacement ratio increases, showing that the Bridge output carries sample-specific information used by the target agent.

We also vary the communication budget by retaining different proportions of the Bridge-output states.
Accuracy decreases for most configurations as fewer states are retained, with different model pairs showing different rates of degradation.
The communication budget therefore controls both processing cost and the amount of sample information available to the target agent.
The new model configuration must therefore balance communication length and task accuracy.
The budget curves and intervention details are provided in Appendix~\ref{sec:budget-details}.

\textbf{Representation analysis.}
Nearest-neighbor retrieval and linear probing both perform above their respective baselines, showing that the representations produced by both processing modes retain sample information.
Although representations of the same sample have low raw similarity across the two modes, linear alignment recovers part of their correspondence.
These results indicate that performance differences arise mainly from differences between representation spaces and from how well the target agent reads the transformed representations, rather than from a complete loss of task information.
This also shows why model replacement requires matching the new model's representation space so that the target agent can use the sample information carried by communication.
Complete retrieval, probing, alignment, and receiver-space results are provided in Appendix~\ref{sec:repr-appendix}.

\section{Conclusion}

This work studies interface reuse in existing latent-communication methods after agent model replacement.
We presented \method{}, which establishes a shared communication protocol through the Bridge and trains the interfaces on its two sides in two stages.
After model replacement, it reuses unchanged and matching interfaces and trains only missing ones.
In the accelerated mode, \method{} avoids autoregressive Bridge generation to reduce communication processing time.
Experiments at two model scales show that \method{} maintains effective communication after agent model replacement and improves communication efficiency.
Ablation and representation analyses confirm the roles of the interfaces, training procedure, and communicated representations in task performance.
\section*{Limitations}

\jiajun{\textbf{Interpretability of continuous representations.}
The analyses in \autoref{sec:exp-analysis} and Appendix~\ref{sec:repr-appendix} examine whether sample information is retained and used, but do not explain the meaning of representation positions.
Task outputs alone may not reveal whether errors arise from source-agent generation, interface mapping, or target-agent interpretation.
More direct tools are needed to explain continuous messages and locate communication errors.}

\jiajun{\textbf{Dependence on a fixed Bridge model.}
Interface composition and reuse in \autoref{sec:interface-reuse} assume that the Bridge model remains fixed, allowing the interface on the unchanged side to be reused after agent model replacement.
If the Bridge model changes, the representation mappings on both sides may need adaptation, and the original interfaces may no longer be directly reusable.
This work therefore addresses agent model replacement with a fixed shared communication component, rather than interface transfer after a Bridge-model update.}

\jiajun{\textbf{Model coverage.}
As described in \autoref{sec:exp-setup}, our experiments cover two model families and two model scales to evaluate interface reuse, task accuracy, and communication efficiency across model configurations.
These configurations include differences in hidden dimension and representation space, but do not cover a wider range of architectures, training approaches, or model-size differences.
Other models may require different representation mappings, so the applicability of the current interface structure to broader model combinations remains to be evaluated.}

\jiajun{\textbf{Multi-agent communication scale.}
\autoref{sec:motivation-problem} explains how individual communication relations form systems with different sizes and topologies, providing a basis for interface reuse across relations.
Our experiments primarily test replacement within a single relation.
Successive agent model replacements and multi-hop communication may introduce interactions and error accumulation, whose effects on overall task performance require system-level evaluation.}

\jiajun{\textbf{Tasks and evaluation.}
We evaluate task accuracy after model replacement through mathematical reasoning and offline action prediction, with data and scoring details in Appendix~\ref{sec:data-scoring-details}.
These evaluations test whether communicated representations support final-answer or next-action generation, but do not cover open-ended tasks or complete interactions with continuing environment feedback.
In particular, offline next-action accuracy on ALFWorld does not directly measure full-task success.
Multi-step decision-making and error recovery after model replacement remain to be evaluated through interactive tasks.}
\section*{Ethical Considerations}

\textbf{Models and data.} Our evaluation uses publicly available pretrained models and the MATH and ALFWorld benchmarks.
MATH contains competition problems, and ALFWorld is a simulated household environment.
This study involves no human participants, new human annotation, or new data collection.
Model weights and datasets remain subject to their original licenses and terms of use.

\textbf{Safety and privacy.} Latent inter-agent messages may be difficult to inspect with text-only moderation tools.
Latent communication should not be treated as privacy protection because states may encode task or user information.
After model replacement, deployed systems should reassess safety and retain human oversight for high-impact actions.
Our offline next-action evaluation on ALFWorld does not establish safety for physical agents.

\textbf{AI use statement.} We used AI tools to assist with drafting, translation, language editing, code editing, and figure preparation.
The authors take responsibility for the accuracy and integrity of the paper, including its claims, results, citations, and final content.

\bibliography{references}

\clearpage

\appendix
\section*{Appendix}
\vspace{0.5em}

\section{RQ1: Task Accuracy}
\label{sec:rq1-details}

\jiajun{This section supplements \autoref{sec:exp-replacement} with replacement operations, result settings, and teacher-forced target loss.}
\jiajun{The task-accuracy results are in \autoref{tab:task-accuracy}.}
\jiajun{\autoref{tab:target-loss} reports teacher-forced target loss, and \autoref{tab:cost} summarizes replacement operations.}

\suppressfloats[t]
\begin{table}[t]
\centering
\small
\setlength{\tabcolsep}{4pt}
\renewcommand{\arraystretch}{1.10}
\begin{tabular*}{\columnwidth}{@{\extracolsep{\fill}}lllrr@{}}
\toprule
Bridge & Source & Target & MATH & ALFWorld \\
\midrule
Qwen  & Qwen  & Qwen  & 1.39 & 1.81 \\
Qwen  & Qwen  & Llama & 1.07 & 1.59 \\
Qwen  & Llama & Qwen  & 1.37 & 1.44 \\
Qwen  & Llama & Llama & 1.08 & 1.52 \\
Llama & Qwen  & Qwen  & 1.21 & 1.33 \\
Llama & Qwen  & Llama & 1.78 & 1.59 \\
Llama & Llama & Qwen  & 1.21 & 1.22 \\
Llama & Llama & Llama & 1.80 & 1.52 \\
\midrule
\multicolumn{3}{l}{Mean} & 1.36 & 1.50 \\
\bottomrule
\end{tabular*}
\caption{Mean teacher-forced target loss for the large-model \jiajun{basic mode}.
Each value uses the same examples, interfaces, Bridge outputs, and target-side inputs as the corresponding large-model accuracy result in \autoref{tab:task-accuracy}.}
\label{tab:target-loss}
\end{table}

\begin{table*}[!tp]
\centering
\small
\renewcommand{\arraystretch}{1.12}
\setlength{\tabcolsep}{5pt}
\begin{tabular}{@{}lllcrr@{}}
\toprule
\jiajun{Replacement case}
& Input interface
& Output interface
& New interfaces
& Small (K)
& Large (K) \\
\midrule
\jiajun{Source: no adaptation}
& Reuse mismatched & Reuse matched & 0 & 0 & 0 \\
\jiajun{Source: interface available}
& Select matched & Reuse matched & 0 & 0 & 0 \\
\jiajun{Target: interface available}
& Reuse matched & Select matched & 0 & 0 & 0 \\
\jiajun{Source: interface missing}
& Train matched & Reuse matched & 1 & 57--131 & 229--262 \\
\jiajun{Target: interface missing}
& Reuse matched & Train matched & 1 & 57--131 & 229--262 \\
\bottomrule
\end{tabular}
\caption{Interface operations \jiajun{after agent model replacement}.
Parameter counts are in thousands for rank-32 interfaces and exclude the frozen agent and Bridge-model backbones.
\jiajun{Stored matching interfaces are reused, and only missing interfaces are trained.}}
\label{tab:cost}
\end{table*}

\subsection{Replacement Settings}
\label{sec:replacement-operations}
\label{sec:rq1-settings}

\jiajun{Replacement comparisons hold the Bridge model and the other agent's model fixed.}
\jiajun{When the source-agent model is replaced, the output interface is reused.}
The system selects an input interface matched to the new source-agent model and current Bridge model, or trains this interface if it is unavailable.
\jiajun{When the target-agent model is replaced, the input interface is reused.}
The system selects an output interface matched to the current Bridge model and new target-agent model, or trains this interface if it is unavailable.
\jiajun{Matching interfaces can also be retrieved from other communication relations in the system.}

\jiajun{\autoref{tab:cost} summarizes these operations and their trainable-parameter requirements.}
\jiajun{The replacement-without-adaptation condition retains an input interface matched to the original source model.}
\jiajun{The stored-interface selection condition uses an interface matched to the replacement model.}
\jiajun{The table reports the number of interface parameters trained for each operation.}

\subsection{Task-Accuracy Results}
\label{sec:rq1-results}

\jiajun{\autoref{tab:task-accuracy} groups the eight configurations by Bridge model and reports the basic and accelerated modes at both model scales.}
\jiajun{The mean row gives an unweighted average across the eight configurations for each task and mode.}
\jiajun{Across tasks, small-model means range from 14.8\% to 27.3\% in the basic mode and from 18.0\% to 36.6\% in the accelerated mode.}
\jiajun{The corresponding large-model means range from 28.0\% to 28.7\% and from 44.8\% to 50.5\%.}
\jiajun{The accelerated mode improves the average accuracy in all four model-scale--task comparisons, while individual small-model configurations can favor the basic mode.}

\jiajun{The large-model MATH results for a Qwen Bridge model and Llama target model use target-token re-embedding of the sequence generated by the Bridge.}
\jiajun{This setting retains autoregressive Bridge generation but changes how its output is supplied to the target model.}
\jiajun{The corresponding ALFWorld results use the hidden-state output mapping.}
\jiajun{The message-replacement and budget experiments use the hidden-state mapping throughout, as specified in \autoref{sec:communication-interventions}.}

\subsection{Teacher-Forced Target Loss}
\label{sec:target-loss-details}

\jiajun{Teacher-forced target loss supplements generated-answer accuracy by measuring the target model's cross-entropy on the reference output.}
\jiajun{The model receives the same fixed instruction and continuous representation as in the corresponding accuracy evaluation.}
The evaluation code first computes one mean answer-token loss for each example and then averages these example-level losses.
\jiajun{The loss is computed only at answer-token positions.}

\jiajun{\autoref{tab:target-loss} covers all eight large-model basic-mode configurations on both tasks using their corresponding accuracy-evaluation inputs.}
\jiajun{The mean losses are 1.36 on MATH and 1.50 on ALFWorld.}
\jiajun{Loss and exact-match accuracy describe different aspects of the output distribution, and the generated answer can remain unchanged as loss decreases.}
\jiajun{\autoref{sec:qualitative-cases} illustrates this distinction with individual examples.}

\begin{figure*}[!t]
  \centering
  \includegraphics[width=\textwidth]{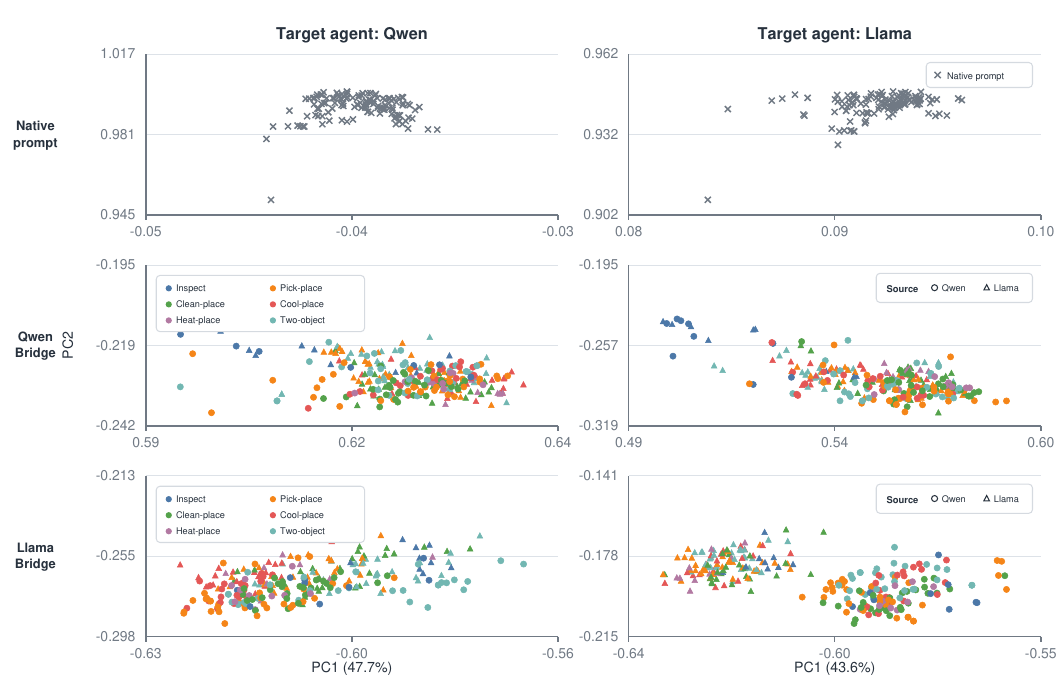}
  \caption{Target-side representation-space analysis on held-out ALFWorld tasks.
  Columns separate target-agent families, while rows show target-native prompt references, Qwen Bridge outputs, and Llama Bridge outputs.
  Colors denote task type and marker shapes denote the source-agent model.
  PCA is fit separately within each target family.
  Each row uses an independent local scale with coordinate values retained on the axes.}
  \label{fig:receiver-space}
\end{figure*}

\section{RQ2: Communication Efficiency}
\label{sec:timing-appendix}

\jiajun{This section supplements \autoref{sec:exp-efficiency} with the timing boundary, hardware settings, and complete latency results.}
\jiajun{The comparison measures communication once the interfaces required by the model configuration are available.}

\jiajun{\autoref{tab:latency} reports the mean across eight configurations for each task and model scale.}
\jiajun{\autoref{tab:latency-detail} gives the individual configuration means.}

\begin{table}[!t]
\centering
\small
\setlength{\tabcolsep}{3pt}
\renewcommand{\arraystretch}{1.10}
\begin{tabular*}{\columnwidth}{@{\extracolsep{\fill}}llrrr@{}}
\toprule
Scale & Task & Before & After & Reduction \\
\midrule
\multicolumn{5}{@{}l}{\jiajun{Interlat (end-to-end)}} \\
-- & ALFWorld & 9{,}190 & 390 & $23.6\times$ \\
\midrule
\multicolumn{5}{@{}l}{\jiajun{\method{} (local Bridge)}} \\
Small & MATH & 1{,}262.53 & 0.334 & $3{,}777\times$ \\
Small & ALFWorld & 559.84 & 0.335 & $1{,}669\times$ \\
Large & MATH & 2{,}256.96 & 0.367 & $6{,}149\times$ \\
Large & ALFWorld & 1{,}181.89 & 0.354 & $3{,}335\times$ \\
\bottomrule
\end{tabular*}
\caption{\jiajun{Mean communication latency (ms).
\method{} compares basic and accelerated modes, averaged over eight configurations.
Interlat compares full and 8-step latent generation~\citep{du2026interlat}.
Reductions use each system's own timing boundary.}}
\label{tab:latency}
\end{table}

\begin{table*}[!t]
\centering
\small
\setlength{\tabcolsep}{4pt}
\renewcommand{\arraystretch}{1.06}
\begin{tabular}{@{}lccc@{\hspace{8pt}}rrrr@{}}
\toprule
Scale & Bridge & Source & Target
& \multicolumn{2}{c}{MATH}
& \multicolumn{2}{c}{ALFWorld} \\
\cmidrule(lr){5-6}\cmidrule(l){7-8}
& & & &
\jiajun{Basic mode}
& \jiajun{Accelerated mode}
& \jiajun{Basic mode}
& \jiajun{Accelerated mode} \\
\midrule
Small & Qwen  & Qwen  & Qwen
& 1545.78 & 0.257 & 651.82 & 0.257 \\
Small & Qwen  & Qwen  & Llama
& 1534.54 & 0.404 & 649.37 & 0.384 \\
Small & Qwen  & Llama & Qwen
& 1508.14 & 0.268 & 655.67 & 0.269 \\
Small & Qwen  & Llama & Llama
& 1499.46 & 0.401 & 572.73 & 0.398 \\
Small & Llama & Qwen  & Qwen
& 987.26 & 0.404 & 573.44 & 0.402 \\
Small & Llama & Qwen  & Llama
& 989.53 & 0.263 & 464.80 & 0.306 \\
Small & Llama & Llama & Qwen
& 1010.12 & 0.405 & 456.70 & 0.413 \\
Small & Llama & Llama & Llama
& 1025.37 & 0.270 & 454.21 & 0.254 \\
\midrule
Large & Qwen  & Qwen  & Qwen
& 2211.68 & 0.290 & 1192.28 & 0.372 \\
Large & Qwen  & Qwen  & Llama
& 2262.02 & 0.592 & 1143.66 & 0.372 \\
Large & Qwen  & Llama & Qwen
& 2208.19 & 0.282 & 1173.31 & 0.284 \\
Large & Qwen  & Llama & Llama
& 2242.59 & 0.432 & 1137.61 & 0.382 \\
Large & Llama & Qwen  & Qwen
& 2260.25 & 0.396 & 1170.86 & 0.397 \\
Large & Llama & Qwen  & Llama
& 2313.33 & 0.261 & 1220.85 & 0.258 \\
Large & Llama & Llama & Qwen
& 2300.87 & 0.419 & 1190.55 & 0.402 \\
Large & Llama & Llama & Llama
& 2256.73 & 0.265 & 1225.97 & 0.367 \\
\bottomrule
\end{tabular}
\caption{Per-configuration \jiajun{Bridge processing} latency in milliseconds.
Each value is the mean over the fixed task test set under the synchronized timing boundary in \autoref{sec:timing-appendix}.}
\label{tab:latency-detail}
\end{table*}

\subsection{Timing Setup}
\label{sec:timing-setup}

\jiajun{The local timer measures input-interface processing, Bridge processing, and output adaptation, ending when the target-side continuous input is ready.}
\jiajun{Device synchronization is applied before and after the measured interval.}
Source-state collection, source generation, target prompt construction, target decoding, sequential model loading, and host-device transfer \jiajun{take place outside the measured interval}.
\jiajun{In the accelerated mode, prompt tokenization and preloading also take place outside this interval.}
All measurements use an NVIDIA GeForce RTX 3090, bfloat16 precision, batch size 1, and 20 warm-up runs.
\jiajun{After warm-up, each configuration is evaluated over the fixed task test set.}

\subsection{Timing Results}
\label{sec:timing-results}

\jiajun{Across the four task-scale comparisons, the accelerated mode reduces mean local communication time from 559.84--2,256.96 ms to 0.334--0.367 ms.}
\jiajun{The reduction ranges from approximately \(1{,}669\times\) to \(6{,}149\times\).}
\jiajun{Once the required mappings are prepared, directly constructing target-side continuous input saves the time spent on autoregressive Bridge generation.}

\jiajun{Interlat's reported timing is included as an external reference.}
\jiajun{Interlat reports end-to-end time, while \method{} reports local Bridge processing time.}
\jiajun{Each reduction is computed within its own system and timing boundary.}
\jiajun{Task-accuracy comparisons for the two modes are reported separately in \autoref{tab:task-accuracy}.}
\section{RQ3: Component and Factor Analysis}
\label{sec:repr-appendix}

\jiajun{This section expands the component and representation analyses in \autoref{sec:exp-analysis}.}
\jiajun{We first examine the interface and training components, then the effects of message content and length, and finally the information retained in different representation spaces.}
\jiajun{Each experiment uses its stated model configuration and data without changing the remaining conditions.}

\begin{figure*}[!tp]
  \centering
  \includegraphics[width=\textwidth]{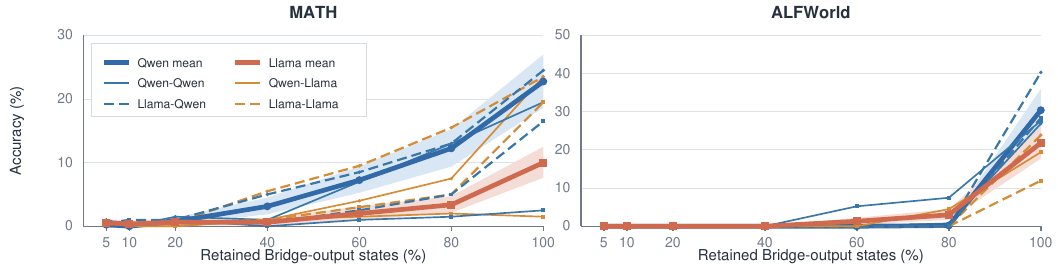}
  \caption{Task accuracy under different communication budgets \jiajun{in the basic mode}.
  \jiajun{We retain different prefixes of the Bridge output before the output interface.}
  \jiajun{Thin curves show source--Bridge pairs, with circles and squares denoting Qwen and Llama targets.}
  \jiajun{Thick curves show target-family means with bootstrap 95\% confidence bands.}}
  \label{fig:bandwidth}
\end{figure*}

\jiajun{\autoref{tab:ablation} and \autoref{tab:representation-summary} summarize the ablation and representation-analysis results.}
\jiajun{\autoref{fig:bandwidth} and \autoref{fig:receiver-space} show the communication-budget results and target-side representation spaces.}

\begin{table*}[!tp]
\centering
\small
\setlength{\tabcolsep}{4pt}
\renewcommand{\arraystretch}{1.18}
\begin{tabular*}{\textwidth}{@{\extracolsep{\fill}}lrrrr@{}}
\toprule
\multirow{2}{*}{Variant}
& \multicolumn{2}{c}{Qwen--Qwen--Llama, MATH}
& \multicolumn{2}{c}{Llama--Qwen--Qwen, ALFWorld} \\
\cmidrule(lr){2-3}\cmidrule(l){4-5}
& Accuracy (\%)
& \jiajun{Parameters (K)}
& Accuracy (\%)
& \jiajun{Parameters (K)} \\
\midrule
\jiajun{\method{} (basic mode)}
& \textbf{11.0} & 151.6 & 22.4 & 188.4 \\
\jiajun{Without} two-stage training
& 2.5 & 151.6 & \textbf{33.6} & 188.4 \\
\jiajun{Without} decoupled output training
& 3.0 & 94.2 & 0.0 & 94.2 \\
\jiajun{Without} input interface
& 2.5 & 94.2 & 0.0 & 94.2 \\
\jiajun{Without} output interface
& 0.5 & 57.3 & 1.5 & 94.2 \\
\jiajun{Without} Bridge-model freezing
& 2.5 & 2{,}314.2 & 22.4 & 3{,}596.3 \\
\bottomrule
\end{tabular*}
\caption{Small-model component and training ablations on two representative configurations.
\jiajun{Configurations are listed as Bridge, source, and target models.}
Learned-parameter counts include all parameters updated by each variant; the final row additionally updates low-rank parameters in the Bridge model.
Bold marks the best accuracy for each task.}
\label{tab:ablation}
\end{table*}

\begin{table}[!tp]
\centering
\small
\setlength{\tabcolsep}{6pt}
\renewcommand{\arraystretch}{1.12}
\begin{tabular}{@{}lrr@{}}
\toprule
Analysis & Qwen & Llama \\
\midrule
Cross-source retrieval & 76.0 & 88.5 \\
Basic-mode probe & \(50.5\pm3.7\) & \(56.5\pm5.6\) \\
Accelerated-mode probe & \(73.0\pm7.3\) & \(71.5\pm4.6\) \\
Cross-mode retrieval & 38.5 & 44.5 \\
\bottomrule
\end{tabular}
\caption{Representation-analysis accuracy (\%) by Bridge model.
Probe values are five-fold means $\pm$ standard deviations.}
\label{tab:representation-summary}
\end{table}

\subsection{Component and Training Ablations}
\label{sec:ablation-details}

\jiajun{The ablations test which parts of interface construction support task accuracy in the small-model basic mode.}
\jiajun{For MATH, the source and Bridge models are Qwen and the target model is Llama.}
\jiajun{For ALFWorld, the source and target models are Qwen and the Bridge model is Llama.}
All variants use seed 0 and keep the training data, evaluation examples, epochs, learning rates, communication budget, and decoding settings fixed.
The complete method uses the corresponding small-model \jiajun{basic-mode results} in \autoref{tab:task-accuracy}.

\jiajun{The five variants change one component or training choice at a time.}
\jiajun{Without two-stage training, the interfaces are jointly optimized.}
\jiajun{Without decoupled output training, the output interface's training states change from separately collected teacher-forced states to Bridge states produced by the current communication process.}
\jiajun{The input-interface and output-interface variants remove the respective trainable mapping.}
\jiajun{Without Bridge-model freezing, rank-32 low-rank parameters in the Bridge model's query and value projections are also updated.}

\jiajun{\autoref{tab:ablation} reports task accuracy and the total number of trainable parameters for each variant.}
\jiajun{Removing either interface or using coupled output training reduces accuracy on both tasks.}
\jiajun{Joint optimization has task-dependent effects, while updating the Bridge model adds more than two million trainable parameters without a consistent accuracy improvement.}
\jiajun{The separate interfaces and training stages therefore support communication while keeping the Bridge model frozen.}

\subsection{Communication Content}
\label{sec:communication-interventions}
\label{sec:message-replacement-details}

\jiajun{Message replacement tests whether the target model uses information specific to the current sample.}
\jiajun{This experiment uses the large models, a communication budget of 128, and the fixed evaluation sets described in \autoref{sec:data-splits}.}
\jiajun{Bridge hidden states are passed through the trained output interface in every configuration, including the Qwen-Bridge-to-Llama-target configurations.}
\jiajun{This keeps the intervention on the same hidden-state path.}

\jiajun{Before the output interface, we replace randomly selected rows of a sample's Bridge output with rows from another sample.}
\jiajun{The replacement sample is selected from the same MATH subject or ALFWorld task type and matched by sequence length.}
\jiajun{Its states are adjusted using row-wise root-mean-square matching.}
\jiajun{Replacement proportions are \(0\), \(0.10\), \(0.25\), \(0.50\), \(0.75\), \(0.90\), and \(1.00\).}
\jiajun{Models, interfaces, and greedy decoding remain fixed.}

\jiajun{\autoref{fig:dose-response} shows the results for perturbation seed 0.}
\jiajun{Target-family means average the four source--Bridge combinations.}
\jiajun{The plotted 95\% confidence bands use 10,000 bootstrap resamples of task identifiers, keeping the configurations for each example together.}
\jiajun{For Qwen targets, macro-average accuracy decreases from 22.7\% to 0.4\% on MATH and from 30.4\% to 7.5\% on ALFWorld.}
\jiajun{For Llama targets, it decreases from 10.0\% to 0.8\% and from 21.8\% to 6.5\%, respectively.}
\jiajun{These changes show that the target output depends on the sample information carried by the message.}
\jiajun{A fully mismatched message and the no-communication condition can produce different results.}

\subsection{Communication Budget}
\label{sec:budget-details}

\jiajun{The budget experiment tests how much of the Bridge-output sequence is needed by the target model.}
\jiajun{It uses the same hidden-state paths, evaluation examples, and trained interfaces as the message-replacement experiment.}
\jiajun{We retain the first \(5\%\), \(10\%\), \(20\%\), \(40\%\), \(60\%\), \(80\%\), or \(100\%\) of the valid Bridge-output states before the output interface.}
\jiajun{The models and decoding settings remain unchanged.}

\jiajun{\autoref{fig:bandwidth} reports the configuration-level curves and target-family means.}
\jiajun{The confidence bands use the same task-level bootstrap procedure as the message-replacement plot.}
\jiajun{Accuracy decreases for most configurations as fewer states are retained, with different Bridge and target model pairs showing different rates of degradation.}
\jiajun{This experiment relates available message length to task accuracy.}
\jiajun{Truncation is applied after Bridge generation, so all retention proportions share the preceding complete autoregressive generation process.}

\subsection{Representation Retrieval, Probing, and Alignment}
\label{sec:retrieval-alignment}

\jiajun{These analyses test whether sample information remains recoverable across source models and processing modes.}
\jiajun{They use the large models and a separate set of 200 MATH algebra examples.}
\jiajun{Each representation sequence is mean-pooled into one vector.}
\jiajun{Cross-source retrieval compares the two source models while holding the Bridge model fixed.}
\jiajun{The correct neighbor is the representation of the same task among 200 candidates, giving a random baseline of 0.5\%.}

\jiajun{The linear probe predicts four clusters constructed from task-text TF-IDF features.}
\jiajun{The reported basic-mode probes use Qwen source states with each Bridge model, while accelerated-mode probes use matching Bridge and target model families.}
\jiajun{The clusters contain 33, 49, 79, and 39 examples, giving a majority-class baseline of 39.5\%.}
\jiajun{The probe standardizes the pooled vectors, reduces them to 64 principal components, and fits logistic regression within five-fold stratified cross-validation.}
\jiajun{For cross-mode alignment, separate 64-dimensional PCA transforms and a ridge mapping are fitted on four folds and evaluated on the held-out fold.}
\jiajun{Retrieval among the 40 held-out candidates has a random baseline of 2.5\%.}

\jiajun{\autoref{tab:representation-summary} reports the retrieval and probing results.}
\jiajun{Both modes predict the text-cluster labels above the majority-class baseline, with higher probe accuracy for the accelerated-mode representations.}
\jiajun{Cross-source retrieval also remains well above chance.}
\jiajun{Although raw cosine similarity across modes is close to zero, linear alignment recovers part of the sample correspondence.}
\jiajun{The results show that differences between representation spaces coexist with recoverable sample information.}

\subsection{Target-Side Representation Space}
\label{sec:receiver-space-details}

\jiajun{This analysis examines how the output interfaces place communicated representations in the target model's input space.}
\jiajun{It uses 134 held-out ALFWorld examples for each of the eight large-model configurations.}
\jiajun{We mean-pool the valid target-side message states and normalize each vector.}
\jiajun{Target-native references are mean-pooled task-token embeddings used only for representation analysis.}

\jiajun{For each target family, PCA is fitted once to the four sets of composed messages and one native reference per task.}
\jiajun{This gives 536 message vectors and 134 reference vectors.}
\jiajun{\autoref{fig:receiver-space} shows the native, Qwen-Bridge, and Llama-Bridge regions under this shared transform, with independently labeled local scales.}

\jiajun{The representations separate mainly by source--Bridge configuration.}
\jiajun{Nearest-neighbor accuracy for identifying the same configuration is 100\%, while same-task accuracy is 0\% for both target families.}
\jiajun{Task-type silhouette scores are \(-0.008\) for Qwen targets and \(-0.006\) for Llama targets.}
\jiajun{The visible geometry therefore mainly reflects representation source and interface differences, with weak separation by task category.}
\section{Additional Related Work}
\label{sec:additional-related-work}

\jiajun{This section expands the discussion in the main Related Work section.}
\jiajun{We first review agent organization and model selection, then latent communication, and finally efficient processing and representation mappings related to \method{}.}

\subsection{Agent Organization and Model Selection}

\jiajun{Research on agent organization explains why a multi-agent system may need to change its models while retaining its task roles and communication relations.}
\jiajun{Role-playing, software-development organizations, reusable conversation frameworks, and evaluation suites study how agents acquire responsibilities and coordinate~\citep{li2023camel,hong2024metagpt,wu2024autogen,zhu2025multiagentbench,chen2026fivews}.}
\jiajun{MasRouter jointly chooses a collaboration mode, role assignment, and language model~\citep{yue2025masrouter}, while sparse and learned topologies change information propagation and computation~\citep{li2024sparse,shen2025topology}.}
\jiajun{ReConcile aggregates diverse answers through confidence-weighted discussion~\citep{chen2024reconcile}.}
\jiajun{Controlled debate studies further show that gains depend on diversity, prompting, topology, and protocol design~\citep{liang2024debate,smit2024mad,wang2024rethinking,zhu2026demystifying}.}
\jiajun{These studies motivate flexible agent configurations, whereas our work examines how latent communication can continue after an agent's model is replaced.}

\subsection{Latent Communication}

\jiajun{Latent-communication methods exchange model representations instead of relying entirely on text.}
\jiajun{CIPHER sends expectations over token embeddings~\citep{pham2024cipher}.}
\jiajun{Other methods communicate activations, token-level state-change trajectories, hidden states, or selected key-value layers~\citep{ramesh2025activations,tang2025sde,du2026interlat,shi2026kvcomm}.}
\jiajun{Recent work also studies joint latent collaboration, cache-to-cache transfer, mixtures of hidden thoughts, learned thought messages, query-conditioned cache exchange, cross-model key-value conversion, and vision-language feature transfer~\citep{zou2025latentmas,fu2025cachetocache,feinashley2025mot,zheng2025thought,kriuk2025qkvcomm,liu2026xkv,liu2026wormhole}.}
\jiajun{These approaches demonstrate communication within their evaluated configurations, but do not establish whether the same interfaces remain usable after agent model replacement.}
\jiajun{Surveys and audits also examine semantic fidelity, causal dependence, and message integrity alongside efficiency~\citep{liu2026beyondtokens,cheng2026causalaudit,brito2026integrity}.}
\jiajun{Unlike work focused on communication within fixed model configurations, \method{} supports agent model replacement by reusing unaffected and existing matching interfaces and training only missing ones.}

\subsection{Efficient Processing}

\jiajun{Related efficiency methods change either the reasoning process or the information exchanged between models.}
\jiajun{Transcript internalization distills a multi-agent interaction into one model~\citep{yi2026latentagents}.}
\jiajun{Continuous and compressed reasoning studies computation within a single model~\citep{hao2025coconut,shen2025codi,cheng2024compressedcot,liu2024deliberation,goyal2024pause,pfau2024dot,chang2026latent}, while prompt and rationale compression shortens text inputs~\citep{jiang2023llmlingua,ren2023sentinel,xia2025tokenskip}.}
\jiajun{In contrast, \method{} retains distinct source and target agents and reduces the processing time of the shared communication component while supporting model replacement.}
\jiajun{It does not merge collaboration into a single model or merely compress text.}

\subsection{Representation Mappings}

\jiajun{Research on representation mappings provides methods for converting representations between models.}
\jiajun{Model stitching studies connections between model representations, while low-rank adaptation reduces the number of trainable parameters~\citep{bansal2021stitching,hu2022lora}.}
\jiajun{Vocabulary adaptation and embedding alignment address related differences between tokenizers~\citep{li2025tokalign,truong2025emo,singh2026byte}.}
\jiajun{Research on representational convergence and representation similarity also informs our analysis of sample correspondence across models~\citep{huh2024platonic,kornblith2019cka}.}
\jiajun{\method{} draws on these representation-mapping methods to implement a shared communication protocol.}
\jiajun{Its focus is on independently matching and reusing interfaces after agent model replacement, rather than only learning mappings between fixed models.}
\section{Implementation Details}
\label{sec:implementation-details}

\jiajun{This section supplements the workflow in \autoref{sec:latbridge} with details of representation handling, interface records, and accelerated-mode processing.}
\jiajun{Training hyperparameters are reported separately in \autoref{sec:training-settings}.}

\subsection{Representation and Input Construction}
\label{sec:representation-implementation}

\jiajun{Representation handling keeps the source states, Bridge states, and target inputs associated with the same task example.}
All states are matched to examples by task identifier.
\jiajun{The input interface receives the source hidden-state sequence after length control by \(\mathcal{P}_L\).}
\jiajun{The communication budget is \(L=128\).}
\jiajun{Valid positions are distinguished from padding by an attention mask.}

For \jiajun{basic-mode} output-interface training, the state collector uses the supervised answer interval under teacher forcing.
\jiajun{During basic-mode evaluation, the Bridge model instead generates a sequence, and the output interface receives the hidden states at its generated positions.}
\jiajun{The target agent receives the resulting continuous input together with its fixed task instruction \(p_t\).}
\jiajun{The source input and Bridge-side task prompt are not separately supplied as text to the target agent.}
\jiajun{The output-mapping variant used in two large-model MATH configurations is described in \autoref{sec:rq1-settings}.}

\subsection{Interface Storage and Retrieval}
\label{sec:interface-records}

\jiajun{Interface records allow a trained mapping to be found and reused for a matching model pair.}
Each stored interface record contains the connected models, interface direction, input and output dimensions, trained parameters, and the processing mode when it affects the output representation.
\jiajun{The source-agent and Bridge models identify an input interface, while the Bridge and target-agent models identify an output interface.}
\jiajun{A stored interface from another communication relation can be reused when these fields match.}
\jiajun{Basic-mode and accelerated-mode output interfaces are distinguished because they are trained on different input representations.}

\subsection{Accelerated-Mode Processing}
\label{sec:accelerated-mode-details}

\jiajun{The accelerated mode constructs target-side continuous input from the Bridge-side task prompt using the prepared vocabulary mapping and output interface.}
\jiajun{The Bridge tokenizer first tokenizes \(p_b\), and the anchor-token suffix used to mark the input position is removed.}
\jiajun{The implementation retains the last \(L\) tokens if the remaining sequence is longer than the communication budget.}
\jiajun{When padding is used, a validity mask \(M\) identifies the positions supplied to the target model with \(p_t\) and \(U_t^{\mathrm{tab}}\).}

\jiajun{When the vocabularies match, Bridge-side token identifiers can be used directly for target embedding lookup.}
\jiajun{Otherwise, the precomputed vocabulary mapping expands each Bridge token into its corresponding target-token sequence.}
\jiajun{The expanded sequence is again limited to the last \(L\) tokens before querying the target embedding table.}
\jiajun{The output interface then transforms the embeddings as specified in \autoref{eq:accelerated-mode}.}

\jiajun{The vocabulary mapping is constructed and cached before evaluation.}
\jiajun{For the timing experiment, tokenization and input preloading are also completed before the measured interval, and embedding lookup is performed on the GPU.}
\jiajun{The complete timing boundary is given in \autoref{sec:timing-appendix}.}

\section{Pilot Study}
\label{sec:pilot-details}

\jiajun{This section provides the settings and reporting details for the pilot study in \autoref{sec:motivation-pilot}.}
\jiajun{The experiment tests whether an interface trained by an existing latent-communication method remains usable after an agent model is replaced.}

\subsection{Experimental Setup}
\label{sec:pilot-setup}

\jiajun{We use Interlat~\citep{du2026interlat} with Qwen2.5-0.5B-Instruct and Llama-3.2-1B-Instruct as the source- and target-agent models.}
\jiajun{For each target model, we compare three conditions.}
\jiajun{The original model combination uses the source model for which the interface was trained.}
\jiajun{The model-replacement condition changes the source model while keeping the target model and original interface fixed.}
\jiajun{The no-communication condition removes the communicated representation and retains only the target agent's fixed instruction.}

\jiajun{The source agent reads the task and its context and produces a continuous representation.}
\jiajun{The target agent receives only this representation and a fixed task instruction, without directly observing the task input.}
\jiajun{We evaluate 200 MATH examples using final-answer exact match and 134 held-out ALFWorld examples using normalized next-action exact match.}
\jiajun{These task scores measure whether the interface passes useful information.}

\jiajun{When the source models have different hidden dimensions, an input projection maps the replacement model's states to the dimension required by the original interface.}
\jiajun{This projection changes the tensor dimension without retraining the original communication interface.}
\jiajun{The three conditions otherwise retain the same target-side task setting.}

\subsection{Results and Reporting}
\label{sec:pilot-results}

\jiajun{\autoref{tab:pilot} reports the mean and standard deviation over three recorded random seeds.}
\jiajun{For a fixed target model, communication gain is the accuracy of a condition minus the no-communication accuracy.}
\jiajun{The retained gain is the replacement condition's gain divided by the original model combination's gain.}

\jiajun{The original combinations provide clear gains on both tasks.}
\jiajun{After replacement, only \(1.5\%\) to \(10\%\) of the original gain remains on MATH, and neither ALFWorld replacement condition retains a positive gain.}
\jiajun{Since the target model and original interface are unchanged, the comparison shows that the original interface is not directly reusable for the new source model.}
\jiajun{This result motivates the separate, reusable interfaces developed in \autoref{sec:latbridge}.}

\section{Data and Evaluation}
\label{sec:data-scoring-details}

\jiajun{This section describes the datasets, evaluation inputs, scoring rules, and resource terms.}
\jiajun{The pilot study and each research-question section specify their own experimental conditions.}
\jiajun{Training settings are given in \autoref{sec:training-settings}.}

\subsection{Datasets and Evaluation Splits}
\label{sec:data-splits}

\jiajun{We use English mathematical problems from MATH~\citep{hendrycks2021math} and English task descriptions and actions from ALFWorld~\citep{shridhar2021alfworld}.}
\jiajun{The large-model interface-training data contain 2,614 MATH algebra and geometry examples and 2,985 ALFWorld examples.}
\jiajun{ALFWorld evaluation uses 134 held-out examples.}
\jiajun{MATH evaluations use 200 examples, with fixed task identifiers specifying each evaluation set.}

\jiajun{The message-replacement and communication-budget experiments use a seed-42 MATH sample with 100 algebra and 100 geometry examples.}
\jiajun{The retrieval, probing, alignment, and qualitative analyses use a separate set of 200 algebra examples selected by task-identifier order.}
\jiajun{Their respective settings are described in \autoref{sec:communication-interventions} and \autoref{sec:retrieval-alignment}.}
\jiajun{These sets serve different analyses and are kept distinct when reporting results.}

\subsection{Task Inputs and Scoring}
\label{sec:scoring-details}

\jiajun{Evaluation preserves the target-side input condition across the compared configurations.}
\jiajun{The source agent receives the task and context, and the Bridge receives the Bridge-side task prompt.}
\jiajun{The target agent receives only its fixed instruction and the continuous input constructed by the Bridge.}
\jiajun{In the no-communication condition, this continuous input is removed.}

MATH scoring extracts a boxed answer when present and applies LaTeX normalization before exact comparison.
\jiajun{The reported small-model evaluation also uses fallback extraction for responses without a boxed answer, checking mathematical expressions, explicit answer phrases, and a final standalone number.}
\jiajun{ALFWorld scoring normalizes the predicted and reference actions before exact comparison.}
\jiajun{Its score measures offline next-action prediction rather than full interactive task success.}

\jiajun{Teacher-forced target loss provides a separate measure of the target model's predictions on the reference output.}
\jiajun{Its computation and results are given in \autoref{sec:target-loss-details}, rather than combined with the task-accuracy metric.}

\subsection{Resource Sources and Licenses}
\label{sec:resource-licenses}

\jiajun{The model and dataset citations identify the original resources, while this subsection records their access sources and license names.}
\jiajun{The MATH loader uses the \texttt{EleutherAI/hendrycks\_math} distribution.}
\jiajun{ALFWorld examples are read from the supervised task-action data used by the evaluation pipeline.}

\jiajun{Qwen2.5-0.5B-Instruct and Qwen2.5-7B are distributed under Apache License 2.0.\footnote{\url{https://huggingface.co/Qwen/Qwen2.5-0.5B-Instruct/blob/main/LICENSE} and \url{https://huggingface.co/Qwen/Qwen2.5-7B/blob/main/LICENSE}.}}
\jiajun{Llama-3.2-1B-Instruct and Llama-3.1-8B-Instruct are subject to their respective Llama 3.2 and Llama 3.1 Community License Agreements and acceptable-use policies.\footnote{\url{https://github.com/meta-llama/llama-models/blob/main/models/llama3_2/LICENSE} and \url{https://github.com/meta-llama/llama-models/blob/main/models/llama3_1/LICENSE}.}}
\jiajun{The recorded configurations load the Llama weights from Unsloth distributions of these models.}
\jiajun{The MATH and ALFWorld project repositories publish MIT licenses for their repository contents.\footnote{\url{https://github.com/hendrycks/math/blob/main/LICENSE} and \url{https://github.com/alfworld/alfworld/blob/master/LICENSE}.}}
\jiajun{These repository licenses should not be taken as a separate grant of rights over all underlying competition problems or third-party resources.}
\jiajun{Use and redistribution remain subject to the terms of the particular model and data distributions.}

\section{Training Settings}
\label{sec:training-settings}
\label{sec:interface-implementation-details}

\jiajun{This section reports the model dimensions and interface-training settings used by the experiments.}
\jiajun{It complements the training objectives in \autoref{sec:interface-construction} and the implementation details in \autoref{sec:implementation-details}.}
\jiajun{Changes made for component ablations are described separately in \autoref{sec:ablation-details}.}

\subsection{Models and Interface Parameters}
\label{sec:model-training-settings}

The small-model experiments use Qwen2.5-0.5B-Instruct with a hidden dimension of 896 and Llama-3.2-1B-Instruct with a hidden dimension of 2,048.
The large-model experiments use Qwen2.5-7B with a hidden dimension of 3,584 and Llama-3.1-8B-Instruct with a hidden dimension of 4,096.
\jiajun{At each scale, selecting source, Bridge, and target models from the two families gives eight configurations.}
\jiajun{The Bridge model remains fixed within each replacement comparison.}

\jiajun{The standard low-rank interface uses rank \(r=32\), scaling coefficient \(\alpha=64\), and dropout 0.1.}
\jiajun{Its trainable matrices follow \autoref{eq:low-rank-interface}.}
\jiajun{Interface parameters are trained while the language-model parameters remain frozen.}
\jiajun{The small-model result records also include output-interface variants and ensembles, so the rank-32 setting describes the standard interface rather than every small-model variant.}
\jiajun{The parameter counts for the evaluated ablations are reported in \autoref{tab:ablation}.}

\subsection{Two-Stage Interface Training}
\label{sec:two-stage-training-settings}

\jiajun{The two stages train the adjacent mappings separately, allowing an existing interface to be reused when the other side changes.}
\jiajun{The input interface is trained for three epochs with a learning rate of \(2\times10^{-4}\).}
\jiajun{The output interface is trained for ten epochs with a learning rate of \(5\times10^{-4}\).}
\jiajun{The standard configurations use AdamW and training seed 0.}
\jiajun{For these configurations, the small-model micro-batch size is 2 with four accumulation steps, while the large-model micro-batch size is 1 with eight accumulation steps.}
\jiajun{Both therefore have an effective batch size of 8.}

\jiajun{Input-interface training supplies mapped source states to the frozen Bridge model and minimizes \autoref{eq:input-interface-loss}.}
\jiajun{Basic-mode output-interface training uses teacher-forced Bridge states from the supervised answer interval and minimizes \autoref{eq:output-interface-loss}.}
\jiajun{States and reference outputs are matched by task identifier.}
\jiajun{After training, each interface is stored using the fields in \autoref{sec:interface-records}.}

\subsection{Accelerated-Mode Output-Interface Training}
\label{sec:accelerated-training-settings}

\jiajun{The accelerated mode requires an output interface trained for the embedding-based input that it will receive during communication.}
\jiajun{The training procedure constructs target embeddings from the Bridge-side task prompt using the vocabulary mapping described in \autoref{sec:accelerated-mode-details}.}
\jiajun{These embeddings replace teacher-forced Bridge states in the target-model cross-entropy objective.}
\jiajun{The large-model experiments use the output-interface settings given above.}
\jiajun{The small-model accelerated-mode report uses rank-32 output interfaces trained for five epochs with AdamW, a learning rate of \(5\times10^{-4}\), and dropout 0.1.}
\jiajun{The resulting interfaces are stored by their Bridge model, target model, and processing mode.}

\section{Qualitative Cases}
\label{sec:qualitative-cases}

\jiajun{This section illustrates how target outputs can differ even when the communicated representations concern the same task.}
\jiajun{The three MATH examples use Qwen2.5-7B as the source, Bridge, and target models and come from the representation-analysis set in \autoref{sec:retrieval-alignment}.}
\jiajun{We retain the sample identifiers to distinguish the cases.}

\subsection{Correct Bridge Answer but Incorrect Target Answer}

\jiajun{In \texttt{MATH\_100}, the task gives \(f(x)=(x+5)/3\) and \(g(x)=1/(f^{-1}(x)+1)\), and asks for \(g(3)\).}
\jiajun{The reference answer is \(1/5\).}
\jiajun{The Bridge model generates the correct answer, but the target agent outputs \(1\) after receiving the basic-mode representation.}
\jiajun{With the accelerated-mode input constructed from the task prompt, the target instead outputs \(1/5\).}
\jiajun{The difference illustrates that a correct answer in the Bridge's generated sequence does not ensure that the target can read it from the transformed hidden states.}

\subsection{An Answer Corrected by the Accelerated Mode}

\jiajun{In \texttt{MATH\_1001}, the task asks for the positive value of \(\sqrt{x^3-2^y}\) when \(x=5\) and \(y=2\).}
\jiajun{The reference answer is \(11\).}
\jiajun{The Bridge's basic-mode generation reaches \(\sqrt{121}=11\), but the target does not produce the correct answer.}
\jiajun{With the accelerated-mode representation, the target outputs \(11\).}
\jiajun{As in the first case, the outcome depends on how the information is presented to the target model.}

\subsection{Lower Loss without a Correct Answer}

\jiajun{In \texttt{MATH\_10}, the task asks for the smallest \(x\) satisfying \(\lvert5x-1\rvert=\lvert3x+2\rvert\).}
\jiajun{The reference answer is \(-1/8\).}
\jiajun{Both modes produce \(1/8\) and miss the required negative solution.}
\jiajun{The accelerated mode nevertheless lowers the teacher-forced target loss.}
\jiajun{This case explains why we report generated-answer accuracy and target loss separately in \autoref{sec:target-loss-details}.}

\end{document}